\documentclass[letterpaper]{article}
\usepackage{proceed2e}
\usepackage[margin=1in]{geometry}

\usepackage{times}
\newcommand{\mat}[1]{\mathbf{\mathrm{#1}}}
\usepackage{natbib}
\usepackage{verbatim} 
\usepackage{hyperref}
\usepackage{amsfonts}
\usepackage{amsmath}
\usepackage{amssymb}
\usepackage{amsthm}
\usepackage{graphicx}
\usepackage{caption}
\usepackage{subcaption}
\usepackage{float}
\usepackage{algorithm}
\usepackage{algorithmic}
\usepackage{dsfont}
\usepackage{sectsty}
\usepackage{titlecaps}
\usepackage{dsfont}
\newcommand{\one}{\mathds{1}}
\newcommand{\krn}{\mat{\Phi}}
\newcommand{\Ocal}{\mathcal{O}}
\newcommand{\Hcal}{\mathcal{H}}

\newcommand{\Xcal}{\mathcal{X}}

\newcommand{\R}{\mathbb{R}}

\title{Incremental kernel PCA and the Nystr\"om method}

\author{Fredrik Hallgren\\
Department of Statistical Science \\
University College London \\
London WC1E 6BT, United Kingdom \\
\texttt{fredrik.hallgren@ucl.ac.uk}
\And
Paul Northrop \\
Department of Statistical Science \\
University College London \\
London WC1E 6BT, United Kingdom \\
\texttt{p.northrop@ucl.ac.uk}}

\begin{document}

\thinmuskip=2mu
\medmuskip=2mu
\thickmuskip=2mu
\maketitle

\sectionfont{\MakeUppercase}
\subsectionfont{\MakeUppercase}

\begin{abstract}
    Incremental versions of batch algorithms are often desired, for increased time efficiency in the streaming data setting, or increased memory efficiency in general. In this paper we present a novel algorithm for incremental kernel PCA, based on rank one updates to the eigendecomposition of the kernel matrix, which is more computationally efficient than comparable existing algorithms. We extend our algorithm to incremental calculation of the Nystr\"om approximation to the kernel matrix, the first such algorithm proposed. Incremental calculation of the Nystr\"om approximation leads to further gains in memory efficiency, and allows for empirical evaluation of when a subset of sufficient size has been obtained.
\end{abstract}

\section{Introduction}

Kernel methods make use of non-linear patterns in data whilst being able to use linear solution methods, through a non-linear transformation of data examples into a feature space where inner products correspond to the application of a kernel function between data examples \citep{hofmann2008kernel}. Many kernel methods have been conceived as the direct application of well-known linear methods in this feature space, occasionally reformulated to be expressed entirely in the form of inner products.

This is the case for kernel PCA, obtained through the application of linear PCA in feature space \citep{scholkopf1998nonlinear} and involving an eigendecomposition of the kernel matrix. It has been shown to outperform linear PCA in a number of applications \citep{chin2007incremental}.

Incremental algorithms, where a solution is updated for additional data examples, are often desirable. If data arrives sequentially in time and a solution is required for each additional data example, more efficient incremental algorithms are often available than repeated application of a batch procedure. Furthermore, incremental algorithms often have a lower memory footprint than their batch counterparts.

In this paper, we propose a novel algorithm for incremental kernel PCA, which accounts for the change in mean in the covariance matrix from each additional data example. It works by writing the expanded mean-adjusted kernel matrix from an additional data point in terms of a number of rank one updates, to which a rank one update algorithm for the eigendecomposition can be applied. We use a rank one update algorithm based on work in \citet{golub1973some} and \citet{bunch1978rank}.

A few previous exact incremental algorithms for kernel PCA have been proposed, some of which are based on the application of an incremental linear PCA method in feature space \citep{kim2005iterative, chin2007incremental, hoegaerts2007efficiently}. Rank one update algorithms for the eigendecomposition have not previously been applied to kernel PCA, to the best of our knowledge. If the mean of the feature vectors is not adjusted, our algorithm corresponds to an incremental procedure for the eigendecomposition of the kernel matrix, which can be more widely applied.

Our algorithm has the same time and memory complexities as existing algorithms for incremental kernel PCA and it is more computationally efficient than the comparable algorithm in \citet{chin2007incremental}, which also allows for a change in mean. Furthermore, it can be considered more flexible, since it is straightforward to apply a different rank one update algorithm to the one we have used, for potentially improved efficiency. Approximate algorithms could also be applied, for example from randomized linear algebra \citep{mahoney2011randomized}.

The usefulness of kernel methods is limited by their large computational requirements in time and memory, which scale in the number of data points, since the dimension of the transformed variables often is very large, or they are not explicitly available, and one therefore must express a solution in terms of transformed data examples. This is particularly true for kernel PCA since it requires an eigendecomposition of the kernel matrix, an expensive operation. As a remedy, various approximate methods have been introduced, such as the Nystr\"om method \citep{williams2001using}, which creates a low-rank approximation to the kernel matrix based on a randomly sampled subset of data examples.

We also extend our algorithm for incremental kernel PCA to incremental calculation of the Nystr\"om approximation to the kernel matrix. We incrementally add data examples to the subset used to create the Nystr\"om approximation to kernel PCA. This allows one to evaluate empirically the accuracy of the Nystr\"om approximation for each added data example. \citet{rudi2015less} presented an incremental updating procedure for the Nystr\"om approximation to kernel ridge regression, based on rank one updates to the Cholesky decomposition. Our proposed incremental procedure can be applied to any kernel method requiring the eigendecomposition or inverse of the kernel matrix. Combining an incremental algorithm with the Nystr\"om method also leads to further improvements in memory efficiency, compared with either method on its own.

\section{Background}

\subsection{Kernel methods}

Kernel methods allow for the application of linear methods to discover non-linear patterns between variables, through a non-linear transformation of data points $\phi(x)$ into a feature space where linear algorithms can be applied \citep{hofmann2008kernel}. They rely on two things. First, the calculation of inner products between transformed data examples through a symmetric positive definite \emph{kernel} $k(x,y)$; second, the expression of a solution linearly in the space of transformed data examples, rather than in the space of transformed variables.

We have a set of $n$ observations $\{x_i\}_{i=1}^n$. Linear methods generally scale in the dimension of the observations. For example, if each $x_i$ is a real vector $x_i = (x_i^{(1)}, x_i^{(2)}, \; \; ..., \; \;  x_i^{(d)})$, a linear method will scale as the number of variables $d$.

Let each $x_i$ be an element from a set $\Xcal$. In general, no further restrictions need to be placed on the set $\Xcal$, which is a great benefit of kernel methods. For example, $\Xcal$ can be a collection of text strings or graphs \citep{lodhi2002text, vishwanathan2010graph}. Let $\Hcal$ be a Hilbert space of real-valued functions on $\Xcal$, with inner product $\langle \, \cdot \, , \, \cdot \, \rangle_{\Hcal} $. If $\Xcal$ is a vector space, then $\Hcal$ is a closed subspace of $\Xcal^*$, the dual space of bounded linear functionals on $\Xcal$.

Consider $\Hcal'$, the dual space of linear functionals on $\Hcal$. For each $x \in \Xcal$ there is an element $\delta_x \in \Hcal'$ such that $\delta_x(f) = f(x)$, termed the evaluation functional. If $\delta_x$ is bounded (i.e. continuous), then by the Riesz representation theorem there is a unique element $g_x \in \Hcal$ such that $\delta_x(f) = \langle g_x, f \rangle_{\Hcal}$ \citep{bollobas1990linear}. If we consider $g_x$ as a function of $x$, say $k(x, \cdot)$, then $k(x, \cdot)$ has the reproducing property, i.e. $\langle k(x, \cdot), f(\cdot) \rangle_{\Hcal} = f(x)$.  Furthermore, by the reproducing property, we have $\langle k(x, \cdot), k(y, \cdot) \rangle_{\Hcal} = k(x,y)$. Then $k(x,y)$ is a symmetric positive definite function by the symmetric positive definite property of the inner product. The function $k(x, \cdot)$ is also often denoted by $\phi(x)$, termed a \emph{feature map}.

The space $\Hcal$ has uncountable dimension, but since every (separable) Hilbert space is isometrically isomorphic to $\ell^2$, the space of square-summable sequences \citep{bollobas1990linear}, each element $\phi(x_i)$ has a representation as a vector $\phi(x_i) = (\phi_1(x_i), \phi_2(x_i), ..., \phi_d(x_i))$ over $\R$  with $\langle \phi(x_i), \phi(x_j) \rangle_{\Hcal} = \sum_{k=1}^d \phi_k(x_i) \phi_k(x_j)$. We call these \emph{feature vectors}. However, this representation is often not known, or $d$ is very large, so it might not be possible to apply a linear method directly on the variables $\phi_1(x), \phi_2(x), ..., \phi_d(x)$. Thanks to the representer theorem \citep{scholkopf2001generalized}, a solution can instead often be expressed in terms of elements in $\Hcal$, as $f(x) = \sum_{i=1}^n \alpha_i k(x_i, x)$ with coefficients $\alpha_i$.

We arrange the feature vectors along the rows of a data matrix $\krn$. The kernel matrix is given by $K := (k(x_i,x_j)) \in \R^{n \times n} = \krn\krn^T$.

\subsection{Kernel PCA}

PCA finds the set of orthogonal linear combinations of variables that maximizes the variance of each linear combination in turn. PCA can be used for dimensionality reduction, in regression and classification problems, and to detect outliers, among other applications \citep{jolliffe2010principal}. The principal components are obtained by calculating the eigendecomposition of the sample covariance matrix $C = \frac{1}{n}X^TX$, for a data matrix of (centred) observations $X$, where each observation occupies a row. This gives the decomposition $C = V \Lambda V^T$ where the columns of $V$ are the directions of maximum variance. The principal components can also be obtained through the related singular value decomposition (SVD).

Assuming centered data, kernel PCA performs the eigendecomposition of the covariance matrix in feature space through \citep{scholkopf1998nonlinear}
\begin{equation*}
\frac{1}{n}\krn^T \krn \mathbf{v} = \lambda \mathbf{v}
\end{equation*}
resulting in the decomposition $\frac{1}{n}\krn^T \krn = V \Sigma V^T$. Henceforth we will ignore the factor $\frac{1}{n}$ and only be concerned with the eigendecomposition of $\krn^T\krn$. Noting that $\mathrm{span} \{\krn^T\} = \mathrm{span} \{ V \}$, we can write $\mathbf{v}$ in terms of an $n$-dimensinal vector $\mathbf{u}$ as $\mathbf{v} = \krn^T \mathbf{u}$. Left-multiplying the eigenvalue equation by $\krn$ we obtain $K \mathbf{u} = \lambda \mathbf{u}$ and the decomposition $K = U \Lambda U^T$.

If the data vectors in feature space are not assumed to be centred, we need to subtract the mean of each variable from $\krn$ and instead calculate the eigendecomposition of
\begin{equation}
    \begin{split}
	K' &= (\krn - \one_{n} \krn )(\krn - \one_n \krn)^T \\
	&= K - \one_{n}K - K\one_{n} + \one_{n}K\one_{n}
    \end{split}
\end{equation}
where $\one_{n}$ is an $n \times n$ matrix for which $(\one_{n})_{i,j} = \frac{1}{n}$, i.e. with every element equal to $\frac{1}{n}$.

\subsection{Incremental kernel PCA}

Incremental algorithms update an existing solution for one or several additional data examples, also referred to as online learning. The goal is that specialized algorithms will achieve greater time or memory performance than repeated application of batch procedures. There are many use cases for incremental versions of batch algorithms, for example when memory capacity is constrained, or when data examples arrive sequentially in time, termed streaming data, and a solution is desired for each additional data example.

A few algorithms for exact incremental kernel PCA have been proposed. The algorithm in \cite{chin2007incremental} is based on the incremental linear PCA algorithm from \citet{lim2004incremental}. The time complexity is  $\Ocal(n^3)$ and the memory complexity $\Ocal(n^2)$. \citet{hoegaerts2007efficiently} write the kernel matrix expanded with an additional data example in terms of two rank one updates, without adjusting for a change in mean, and hence propose an algorithm to update a subset of $m$ dominant eigenvalues and corresponding eigenvectors. If the algorithm is applied to update all eigenpairs, the complexities in time and memory are $\Ocal(n^3)$ and $\Ocal(n^2)$, respectively.

Iterative algorithms produce a sequence of improving approximate solutions that converges to the exact solution as the number of steps increases \citep{golub1983matrix}. An iterative algorithm can often be made to operate efficiently in an incremental fashion, by expanding the data set with additional data examples and restarting the iterative procedure. An example of an iterative method for kernel PCA that can be made to operate incrementally is the kernel Hebbian algorithm \citep{kim2005iterative}, based on the generalized Hebbian algorithm \citep{oja1982simplified} applied in feature space.

Various \emph{approximations} to incremental kernel PCA have also been proposed. See for example \citet{tokumoto2011fast} or \citet{sheikholeslami2015kernel}. Since we present an exact algorithm for incremental kernel PCA, we will not describe these or similar works further.

\subsection{The Nystr\"om method}

The Nystr\"om method \citep{williams2001using} randomly samples $m$ data examples from the full dataset, often uniformly, and calculates a low-rank approximation $\tilde{K}$ to the full kernel matrix through
\begin{equation*}
    \tilde{K} = K_{n,m} K^{-1}_{m,m} K_{m,n}
\end{equation*}
where $K_{n,m}$ is an $n \times m$ matrix obtained by choosing $m$ columns from the original matrix $K$, $K_{m,n}$ is its transpose and $K_{m,m}$ contains the intersection of the same $m$ columns and rows.

\section{Kernel PCA through \\ rank one updates}

\thinmuskip=0.5mu
\medmuskip=0.5mu
\thickmuskip=0.5mu

In this section we present an algorithm for incremental kernel PCA based on rank one updates to the eigendecomposition of the kernel matrix $K$, or the mean-adjusted kernel matrix $K'$. Any incremental algorithm for the eigendecomposition of the kernel matrix $K$ can be applied where the explicit or implicit inverse of the same is required, such as kernel regression and kernel SVM. Various methods other than kernel PCA are also based on the eigendecomposition of the kernel matrix, such as kernel FDA \citep{mika1999fisher}. Even when more efficient solution methods are available, access to the eigendecomposition can be highly useful for statistical regularization or controlling numerical stability.

In contrast to the covariance matrix in linear PCA, the kernel matrix expands in size for each additional data point, which needs to be taken into account, and the effect on the eigensystem determined. We write the kernel matrix $K'_{m+1,m+1}$ created with $m+1$ data examples in terms of an expansion and a sequence of symmetric rank one updates to the kernel matrix $K'_{m,m}$, and apply a rank one update algorithm to the eigendecomposition of $K'_{m,m}$ to obtain the eigendecomposition of $K'_{m+1,m+1}$.

A number of algorithms have been suggested to perform rank one modification to the symmetric eigenproblem. \citet{golub1973some} presented a procedure to determine the eigenvalues of a diagonal matrix updated through a rank one perturbation. \citet{bunch1978rank} extended the results to the determination of both eigenvalues and eigenvectors of an arbitrary perturbed matrix, including an improved procedure to determine the eigenvalues. Stability issues in the calculation of the eigenvectors, including loss of numerical orthogonality, later motivated several improvements \citep{dongarra1987fully, sorensen1991orthogonality, gu1994stable}. Alternatively, one could potentially employ update algorithms for the singular value decomposition, such as the algorithm suggested in \cite{brand2006fast} for the thin singular value decomposition.

We use the rank one update algorithm for eigenvalues from \cite{golub1973some} and the determine the eigenvectors according to \cite{bunch1978rank}. In the experiments our approach seems to be sufficiently stable and accurate for most use cases. We assume throughout that the kernel matrix remains non-singular after each update.

Our algorithm has the same time and memory complexities as competing methods. The algorithm most comparable to ours is the one in \citet{chin2007incremental}, which also accounts for a change in mean. If one additional data example is added incrementally, and all eigenpairs are retained, it requires the eigendecomposition of an $m + 2 \times m + 2$ matrix, the eigendecomposition of the $m \times m$ unadjusted kernel matrix, and a multiplication of two $m \times m$ matrices at each step. Since a multiplication of two $m \times m$ matrices requires $2m^3$ flops, and the state-of-the-art QR algorithm for the symmetric eigenproblem about $9m^3$ flops \citep{golub1983matrix}, the algorithm thus requires $20m^3$ flops to the $\Ocal(m^3)$ factor. Our proposed algorithm requires $8m^3$ flops to the $\Ocal(m^3)$ factor if the mean is adjusted, and $4m^3$ flops otherwise, from one multiplication of two $m + 1 \times m + 1$ matrices for each rank one update. Our algorithm is thus more than twice as efficient.

\subsection{Rank one update procedure}

If we know the eigendecomposition of $K'_{m,m} = U_m \Lambda_m U_m^T$ and write $K'_{m+1,m+1}$ in terms of an expansion and number of symmetric rank one updates to $K'_{m,m}$, we can then apply a rank one update algorithm to obtain the eigendecomposition of $K'_{m+1,m+1} = U_{m+1} \Lambda_{m+1} U_{m+1}^T$.

\subsubsection{Zero-mean data} \label{sec:zeromean}

If we assume that the data examples have zero mean in feature space, then the mean does not need to be updated for previous data points and $K_{m,m}$ only needs to be expanded with an additional row and column. In this case we can devise a rank one update procedure from $K_{m,m}$ to $K_{m+1,m+1}$ in two steps. We denote $k_{i,j} = k(x_i, x_j)$ and $a \; \; \; = \; \; \; [ \; k_{1,m+1} \; \; \; \; \; \; \; \; \; \; \; \; k_{2,m+1} \; \; \; \; \; \; \; \; \; \; \cdots \; \; \; \; \; \; \; \; \; \; \; \;k_{m,m+1} \; ]^T$, i.e. a column vector with elements $k_{1,m+1}, \; \; \; \; \; \; k_{2,m+1},\; \; \; \; \; \; ..., \; \; \; \; \; \;k_{m,m+1}$ and let
\begin{equation*}
	v_1  = [ \; a^T \; \; \; \; \; \; \; \;  \; \; \; \; \; \;\frac{1}{2} k_{m+1,m+1} \;]^T
\end{equation*}
    \begin{equation*}
    \begin{split}
	v_2 &= [ \; a^T \; \; \; \; \; \; \; \;  \; \; \; \; \; \;\frac{1}{4} k_{m+1,m+1} \; ]^T \\
	\sigma &= 4/k_{m+1,m+1}
    \end{split}
\end{equation*}
Then we have
\begin{equation} \label{eq:k0update}
    \begin{split}
	&K_{m+1,m+1} = \\
	&= \begin{bmatrix}
    K_{m,m} & \mathbf{0}_m \\
    \mathbf{0}_m^T & \frac{1}{4} k_{m+1,m+1}
\end{bmatrix}
    + \sigma v_1 v_1^T - \sigma v_2 v_2^T \\
	&:= K^0_{m,m} + \sigma v_1 v_1^T - \sigma v_2 v_2^T
    \end{split}
\end{equation}
corresponding to an expansion of $K_{m,m}$ to $K^0_{m,m}$ and two rank one updates, where $\mathbf{0}_m$ is a column vector of zeros. Compared to the eigensystem of $K_{m,m}$, $K^0_{m,m}$ will have an additional eigenvalue $\lambda_{m+1} = \frac{1}{4} k_{m+1,m+1}$ and corresponding eigenvector $u_{m+1} = [ \; 0 \; \; \; \; \; \; \; \; \; \; \; \; 0 \; \; \; \; \; \; \; \; \; \; \; \; \cdots \; \; \; \; \; \; \; \; \; \; \; \; 0 \; \; \; \; \; \; \; \; \; \; \; \; 1 \;]^T$. The matrix $K^0_{m,m}$ is symmetric positive definite (SPSD), since all eigenvalues are positive. It will remain SPSD after the first update, since it is a sum of two SPSD matrices, as $v_1v_1^T$ is a Gram matrix, if each element is instead seen as a separate vector. The resulting matrix after the second update will be SPSD since this holds for $K_{m+1,m+1}$.  The algorithm for one updating iteration is described in Algorithm 1, given a function $\mathrm{rankoneupdate}(\sigma, v, L, U)$ that updates the eigenvalues $L$ and eigenvectors $U$ from a rank one additive perturbation $\sigma v v^T$.

\begin{algorithm}
    \label{al:kernelupdate1}
    \caption{Incremental eigendecomposition of kernel matrix}
    \begin{algorithmic}[1]
	\small
	\REQUIRE Dataset $\{x_i\}_{i=1}^{m+1}$; row vector of eigenvalues $L$ and matrix of eigenvectors $U$ of $K_{m,m}$; kernel function $k(\cdot,\cdot)$
	\ENSURE Eigenvalues $L$ and eigenvectors $U$ of $K_{m+1,m+1}$
	  \STATE $L \leftarrow [L \; \; \; \; \; \; \; \; \; \; \; \; k_{m+1,m+1} / 4]$
	  \STATE $U \leftarrow \begin{bmatrix} U & 0 \\ 0 & k_{m+1,m+1} / 4 \end{bmatrix} $
	  \STATE sigma $\leftarrow 4 / k_{m+1,m+1}$
	  \STATE $k1 \leftarrow [k_{1,m+1} \; \; \; \; \; \; \; \; \; \; \; \; k_{2,m+1} \; \; \; \; \; \; \; \; \; \; \; \; ... \; \; \; \; \; \; \; \; \; \; \; \; k_{m+1,m+1}/2]$
	  \STATE $k0 \leftarrow [k_{1,m+1} \; \; \; \; \; \; \; \; \; \; \; \; k_{2,m+1} \; \; \; \; \; \; \; \; \; \; \; \; ... \; \; \; \; \; \; \; \; \; \; \; \; k_{m+1,m+1}/4]$
	  \STATE $L, U \leftarrow \mathrm{rankoneupdate(sigma}, \; \; \; \; \; \;k1, \; \; \; \; \; \;L, \; \; \; \; \; \;U \mathrm{)}$
	  \STATE $L, U \leftarrow \mathrm{rankoneupdate(-sigma}, \; \; \; \; \; \;k0, \; \; \; \; \; \;L, \; \; \; \; \; \;U \mathrm{)}$
    \end{algorithmic}
\end{algorithm}

If we limit ourselves to kernel functions for which $k(x,x)$ is constant, without loss of generality we can set $k(x,x) = 1$  and the above expression simplifies.

\subsubsection{Mean-adjusted data}

To construct a rank one update procedure from $K'_{m,m}$ to $K'_{m+1,m+1}$, all the elements of $K'_{m,m}$ need to be adjusted in addition to the expansion with another row and column. We first devise two rank one updates that adjust the mean of $K'_{m,m}$ to account for the additonal data example. We then expand the resulting matrix and perform symmetric updates to set the last row and column to the required values, similarly to (\ref{eq:k0update}).

Recall that when taking the mean into account, one performs an eigendecomposition of the adjusted kernel matrix $K' = K - \one_n K + K \one_n - \one_n K \one_n$. The elements of $K'_{m,m}$ can thus be adjusted through the following formula
\begin{equation*}
    \begin{split}
	&K''_{m,m} := \; \; (K'_{m+1,m+1})_{1:m,1:m} \\
	&= \; \; K'_{m,m} + \one_m K_{m,m} + K_{m,m} \one_m - \one_m K_{m,m} \one_m \\
	   &+ (-\one_{m+1} K_{m+1,m+1} - K_{m+1,m+1} \one_{m+1} \\
	   &+ \one_{m+1} K_{m+1,m+1}\one_{m+1})_{1:m,1:m}
    \end{split}
\end{equation*}
where $( \; \; \; \; \; \; \cdot \; \; \; \; \; \;)_{1:m,1:m}$ denotes the first $m$ rows and columns of a matrix. The latter six terms are all rank one matrices. The matrices $\one_m K_{m,m}$ and $-(\one_{m+1} K_{m+1,m+1})_{1:m,1:m}$ are constant along the columns, and hence their sum, and similarly for the rows of $K_{m,m} \one_m - ( K_{m+1,m+1}\one_{m+1})_{1:m,1:m}$. The matrix $\one_m K_{m,m}\one_m$ has constant entries, equal to the sum of all elements of $K_{m,m}$ multiplied by a factor $1/m^2$, and similarly for $(\one_{m+1} K_{m+1,m+1}\one_{m+1})_{1:m,1:m}$. Consequently, all terms can be written as two rank one updates. We have
\begin{equation*}
    \begin{split}
	\one_m K_{m,m} &- (\one_{m+1} K_{m+1,m+1})_{1:m,1:m} \\
	&= \frac{1}{m+1} (\one_m K_{m,m} -  \mathbf{1}_m a^T) \\
	K_{m,m} \one_m  &- ( K_{m+1,m+1}\one_{m+1})_{1:m,1:m} \\
	&=  \frac{1}{m+1} (K_{m,m}\one_m - a \mathbf{1}_m^T)
    \end{split}
\end{equation*}
with $a$ as in section \ref{sec:zeromean} above and where $\mathbf{1}_m$ is a column vector of ones. Since $K_{m,m}$ is symmetric for all $m$, we have $\one_m K_{m,m} = \; \; \;  (K_{m,m}\one_m)^T$ and $(\one_{m+1} K_{m+1,m+1})_{1:m,1:m} = \; \; \; (K_{m+1,m+1}\one_{m+1})_{1:m,1:m}^T$, and can set
\begin{equation*}
    \begin{split}
	u &= \frac{1}{m(m+1)} K_{m,m} \mathbf{1}_m - \frac{1}{m+1} a + \frac{1}{2} C \mathbf{1}_m \\
	C &= - \frac{1}{m^2} \Sigma_m + \frac{1}{(m+1)^2} \Sigma_{m+1}
    \end{split}
\end{equation*}
 where we have denoted $\Sigma_m = \mathbf{1}_m^T K_{m,m} \mathbf{1}_m$, the sum of all elements of $K_{m,m}$, to obtain
\begin{equation*}
    \begin{split}
    &K''_{m,m} = K'_{m,m} + \mathbf{1}_m u^T + u \mathbf{1}_m^T \\
	&=K'_{m,m} + \frac{1}{2}(\mathbf{1}_m + u)(\mathbf{1}_m + u)^T
- \frac{1}{2}(\mathbf{1}_m - u)(\mathbf{1}_m - u)^T
    \end{split}
\end{equation*}
which is two symmetric rank one updates to $K'_{m,m}$. $\Sigma_m$ and $K_{m,m}\mathbf{1}_m$ can easily be updated between iterations like so
\begin{equation*}
    \begin{split}
	\Sigma_{m+1} &= \Sigma_m + 2a^T\mathbf{1}_m + k_{m+1,m+1} \\
	K_{m+1,m+1} \mathbf{1}_{m+1} &= [ K_{m,m}\mathbf{1}_m + a \; ; \; \; \; \; \; \; a^T \mathbf{1}_m + k_{m+1,m+1} ]
    \end{split}
\end{equation*}
where $[ \; \; \; b \; \; \; ;  \; \; \; \; \; \; c \; \; \; ]$ denotes a column vector $b$ expanded with an additional element $c$. We now expand $K''_{m,m}$ to $K'_{m+1,m+1}$, analogously to (\ref{eq:k0update}), but taking the adjusted mean into account. The required last row and column is given by
\begin{equation*}
    \begin{split}
	v \; \; \; \; \; \; := \; \; \; \; \; \; k - \frac{1}{m+1} ( \mathbf{1}_{m+1} \mathbf{1}^T_{m+1} k &+ K_{m+1,m+1} \mathbf{1}_{m+1} \\
	 &- \frac{1}{m+1} \Sigma_{m+1} \mathbf{1}_{m+1})
    \end{split}
\end{equation*}
with $k = [ \; a^T \; \; \; \; \; \; \; \; \; \; \;  k(x_{m+1},x_{m+1}) \; ]^T$. If we let
\begin{equation*}
    \begin{split}
	v_1 &= [(v)_{1:m} \; ; \; \; \; \; \; \; \frac{1}{2} (v)_{m+1}] \\
	v_2 &= [(v)_{1:m} \; ; \; \; \; \; \; \;  \frac{1}{4} (v)_{m+1}] \\
	\sigma &= 4/(v)_{m+1}
    \end{split}
\end{equation*}
where $(v)_{1:m}$ is a vector of the first $m$ elements of $v$, and $(v)_{m+1}$ is its last element, we have
\begin{equation} \label{eq:kprimeupdate}
    \begin{split}
	K'_{m+1,m+1} &=
	\begin{bmatrix}
    K''_{m,m} & \mathbf{0}_m \\
		\mathbf{0}_m^T & \frac{1}{4} (v)_{m+1}
\end{bmatrix}
    + \sigma v_1 v_1^T - \sigma v_2 v_2^T \\
	:&= K^0_{m,m} + \sigma v_1 v_1^T - \sigma v_2 v_2^T
    \end{split}
\end{equation}
We have thus devised a procedure to update $K'_{m,m}$ to $K'_{m+1,m+1}$ using four symmetric rank one updates, for which a rank one eigendecomposition update algorithm can be applied. The full procedure is described in Algorithm 2. Note that the matrix $K'_{m,m}$ or its expansion do not need to be kept in memory. The procedure is linear in time and memory, since all constituent quantities are updated incrementally.

\thinmuskip=1mu
\medmuskip=1mu
\thickmuskip=1mu
\begin{algorithm}
    \label{al:kernelupdate2}
    \caption{Incremental eigendecomposition of \\ adjusted kernel matrix}
    \begin{algorithmic}[1]
	\small
	\REQUIRE Dataset $\{x_i\}_{i=1}^{m+1}$; row vector of eigenvalues $L$ and matrix of eigenvectors $U$ of $K_{m,m}$; kernel function $k(\cdot,\cdot)$; sum of all elements of $K_{m,m}$, denoted $S$; sum of rows of $K_{m,m}$, i.e. $K_{m,m} \mathbf{1}_m$, denoted $K1$
	\ENSURE Eigenvalues $L$ and eigenvectors $U$ of $K_{m+1,m+1}$
	  \STATE $a \leftarrow [k_{1,m+1} \; \; \; k_{2,m+1} \; \; \; ... \; \; \; k_{m,m+1}]$
	  \STATE $S2 \leftarrow S + 2 * \mathrm{sum(} a \mathrm{)} + k_{m+1,m+1}$
	  \STATE $C \leftarrow -S / m^2 + S2 / (m+1)^2$
	  \STATE $u \leftarrow K1 / (m*(m+1))^2 - a / (m+1) + 0.5 * C * \mathrm{ones(}m\mathrm{)}$
	  \STATE $L, U \leftarrow \mathrm{rankoneupdate}(0.5,\;\;\;\;\;\; 1+u, \;\;\;\;\;\;L, \;\;\;\;\;\;U)$
	  \STATE $L, U \leftarrow \mathrm{rankoneupdate}({-0.5}, \;\;\;\;\;\;1-u, \;\;\;\;\;\;L, \;\;\;\;\;\;U )$
	  \STATE $K1 \leftarrow [K1 + a \; \; \; \; \; \; \; \; \; \; \; \; \mathrm{sum(}a\mathrm{)} + k]$
	  \STATE $S \leftarrow S2$
	  \STATE $m \leftarrow m+1$
	  \STATE $v \leftarrow k - (\mathrm{ones(}m\mathrm{)} * (\mathrm{sum(}a\mathrm{)} + k) + K1 - S/m)/m$
	  \STATE $v0 \leftarrow v[m]$
	  \STATE $v \leftarrow v[1:m-1]$
	  \STATE $L \leftarrow [L \; \; \; \; \; \; \; \; \;v0 / 4]$
	  \STATE $U \leftarrow \begin{bmatrix} U & 0 \\ 0 & v0/4 \end{bmatrix} $
	  \STATE $\mathrm{sigma} \leftarrow 4/v0$
	  \STATE $v1 \leftarrow [v \; \; \; \;\;\;\;\;\;v0/2]$
	  \STATE $v2 \leftarrow [v \; \; \; \;\;\;\;\;\;v0/4]$
	  \STATE $L, U \leftarrow \mathrm{rankoneupdate(sigma},\;\;\;\;\;\; v1, \;\;\;\;\;\;L, \;\;\;\;\;\;U \mathrm{)}$
	  \STATE $L, U \leftarrow \mathrm{rankoneupdate({-sigma}}, \;\;\;\;\;\;v2, \;\;\;\;\;\;L, \;\;\;\;\;\;U \mathrm{)}$
    \end{algorithmic}
\end{algorithm}

\subsection{Update algorithm for the \\ eigendecomposition}

Here we describe an algorithm for updating the eigendecomposition after a rank one perturbation. Suppose we know the eigendecomposition of a symmetric matrix $A = U \Lambda U^T$. Let
\begin{equation*}
    B = U \Lambda U^T + \sigma v v^T = U(\Lambda + \sigma z z^T)U^T
\end{equation*}
where $z = U^T v$, and look for the eigendecomposition of $\tilde{B} = \Lambda + \sigma z z^T := \tilde{U}\tilde{\Lambda}\tilde{U}^T$ \citep{bunch1978rank}. Then the eigendecomposition of $B$ is given by $U\tilde{U} \tilde{\Lambda} \tilde{U}^T U^T$ with unchanged eigenvalues and eigenvectors $U^B := U\tilde{U}$, since the product of two orthogonal matrices is orthogonal and since the eigendecomposition is unique, provided all eigenvalues are distinct.

The eigenvalues of $\tilde{B}$ can be calculated in $\Ocal(n^2)$ time by finding the roots of the secular equation \citep{golub1973some}
\begin{equation} \label{eq:eigvalupd}
    \omega(\tilde{\lambda}) := 1  + \sigma \sum_{i=1}^n \frac{z_i^2}{\lambda_i - \tilde{\lambda}}
\end{equation}
The eigenvalues of the modified system are subject to the following bounds
\begin{equation} \label{eq:eigbounds}
    \begin{aligned}[c]
    & \lambda_i \le \tilde{\lambda}_i \le \lambda_{i+1}  \\
    & \lambda_n \le \tilde{\lambda}_n \le \lambda_n + \sigma z^T z \\
    & \lambda_{i-1} \le \tilde{\lambda}_i \le \lambda_i \\
    & \lambda_1 + \sigma z^T z \le \tilde{\lambda}_1 \le \lambda_1
    \end{aligned}
    \hspace{10pt}
    \begin{aligned}[c]
	i = 1, 2, ..., n-1, \; \; \; \; \; \; &\sigma > 0 \\
	 &\sigma > 0 \\
	i = 2, 3, ..., n, \; \; \; \; \; \; &\sigma < 0 \\
	 &\sigma < 0
    \end{aligned}
\end{equation}
which can be used to supply initial guesses for the root finding algorithm. Note that after expanding the eigensystem, as described above, the eigenpairs need to be reordered for the bounds to be valid.

Once the updated eigenvalues have been calculated the eigenvectors of the perturbed matrix $B$ are given by \citep{bunch1978rank}
\begin{equation} \label{eq:eigvecupd}
    u^B_i = \frac{U D_i^{-1}z}{\| D_i^{-1}z \|}
\end{equation}
where $D_i := \Lambda - \tilde{\lambda}_i I$. Since $U$ and $D_i^{-1}$ are $m \times m$ and $D_i$ is diagonal the denominator is $\Ocal(m)$ and the numerator is $\Ocal(m^2)$, leading to $\Ocal(m^3)$ time complexity to update all eigenvectors. The number of flops for the full procedure is $2n^3 + \Ocal(n^2)$. Equation (\ref{eq:eigvecupd}) requires the creation of an additional $n \times n$ matrix, hence the full procedure is quadratic in memory.

\begin{figure*}
    \begin{subfigure}[b]{0.5\textwidth}
	\includegraphics[width=3.2in]{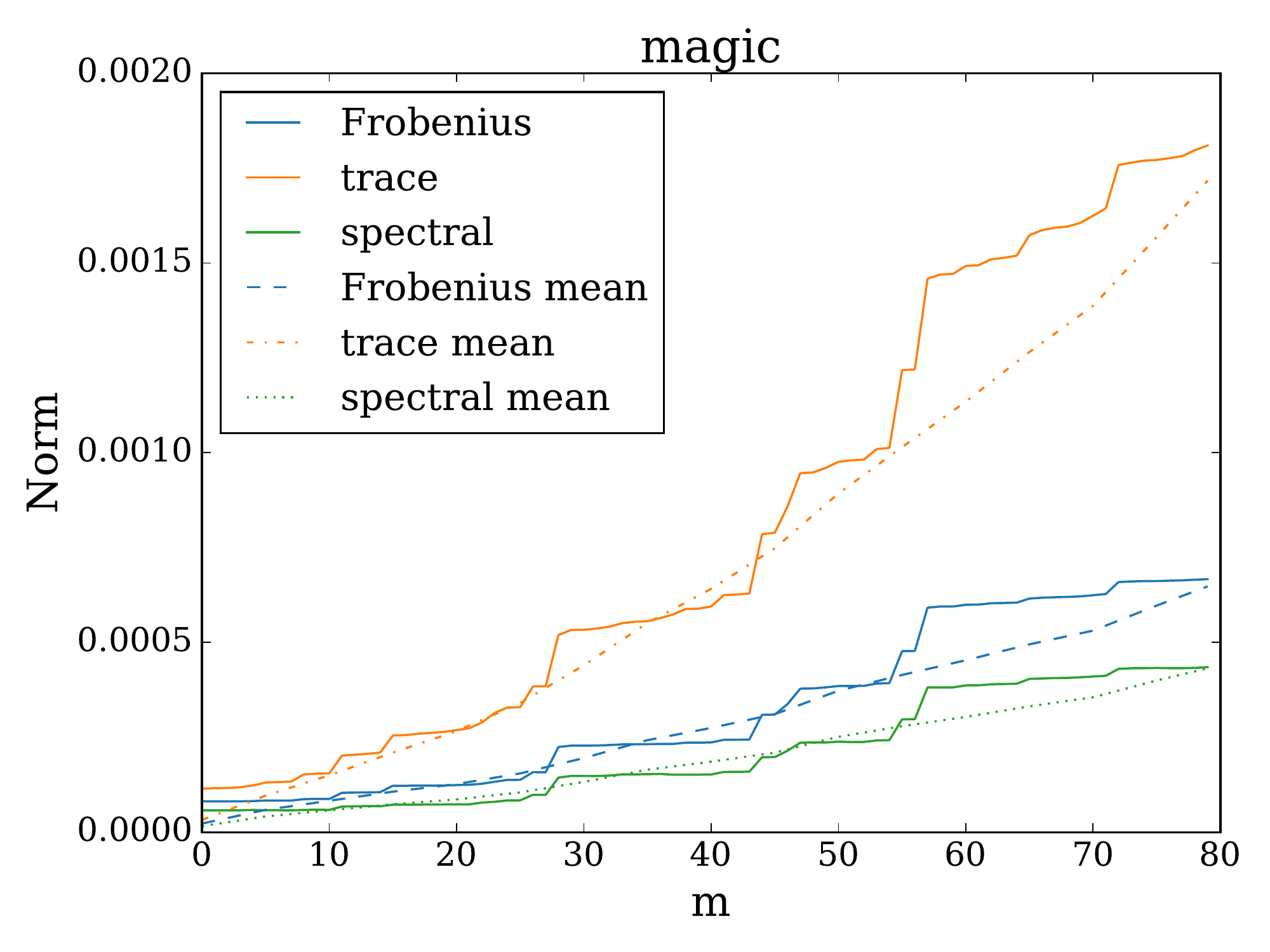}
    \end{subfigure}
    ~
    \begin{subfigure}[b]{0.3\textwidth}
	\includegraphics[width=3.2in]{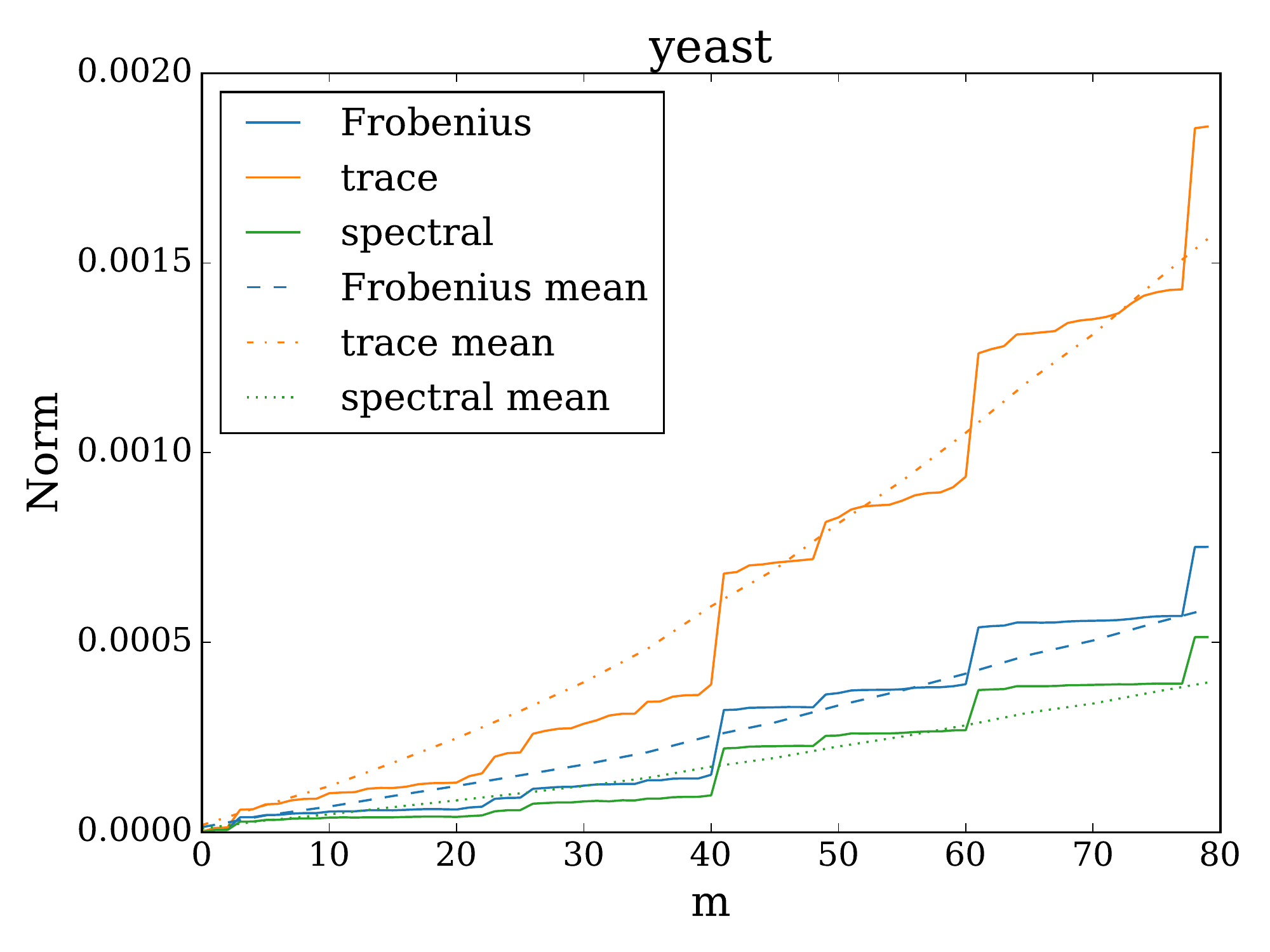}
    \end{subfigure}
    \caption{Difference between batch and incremental calculation of $K'$ of size $20 + m$ for the two datasets.}
	\label{fig:ikpca}
\end{figure*}

\vspace{2pt}
\section{Incremental Nystr\"om}

In this section we extend our proposed algorithm to incremental calculation of the Nystr\"om approximation to the kernel matrix. Having access to an incremental procedure for the Nystr\"om method can be highly useful. Different sizes of subsets used in the approximation can efficiently be evaluated, to determine a suitable size for the problem at hand or for empirical investigation of the characteristics of the Nystr\"om method for subsets of different sizes. For very large datasets, the combination of the Nystr\"om method with incremental calculation results in further gains in memory efficiency.

\citet{rudi2015less} previously proposed an incremental algorithm for the Nystr\"om approximation applied to kernel ridge regression, based on rank one updates to the Cholesky decomposition. Our proposed procedure can be seen as a generalization of their work. To the best of our knowledge, it is the first incremental algorithm for calculation of the full Nystr\"om approximation to the kernel matrix.

Given the eigenvalues $\Lambda$ and eigenvectors $U$ of the matrix $K_{m,m}$, the corresponding approximate eigenvalues and eigenvectors of $K$ are given by \citep{williams2001using}
\begin{equation} \label{eq:eigscale}
    \begin{split}
	& \Lambda^{nys} := \frac{n}{m} \Lambda \\
	& U^{nys} := \sqrt{\frac{m}{n}} K_{n,m} U \Lambda^{-1}
    \end{split}
\end{equation}

To obtain an incremental procedure for $\tilde{K} = U^{nys}\Lambda^{nys}U^{nysT}$, calculate $U$ and $\Lambda$ incrementally using Algorithm~(2), then at each iteration add an extra column to $K_{n,m}$ corresponding to the additional data example, and calculate the rescaling (\ref{eq:eigscale}).
The rescaling has $\Ocal(m^2n)$ time complexity from the matrix product in (\ref{eq:eigscale}).

Note that the proposed incremental calculation of the Nystr\"om approximation exactly reproduces batch computation at each $m$, save for numerical differences. The accuracy of the Nystr\"om approximation has been extensively studied, including comparisons with other methods \citep{gittens2016revisiting, yang2012nystrom}.

\begin{figure*}
    \begin{subfigure} [b]{0.5\textwidth}
	\includegraphics[width=3.2in]{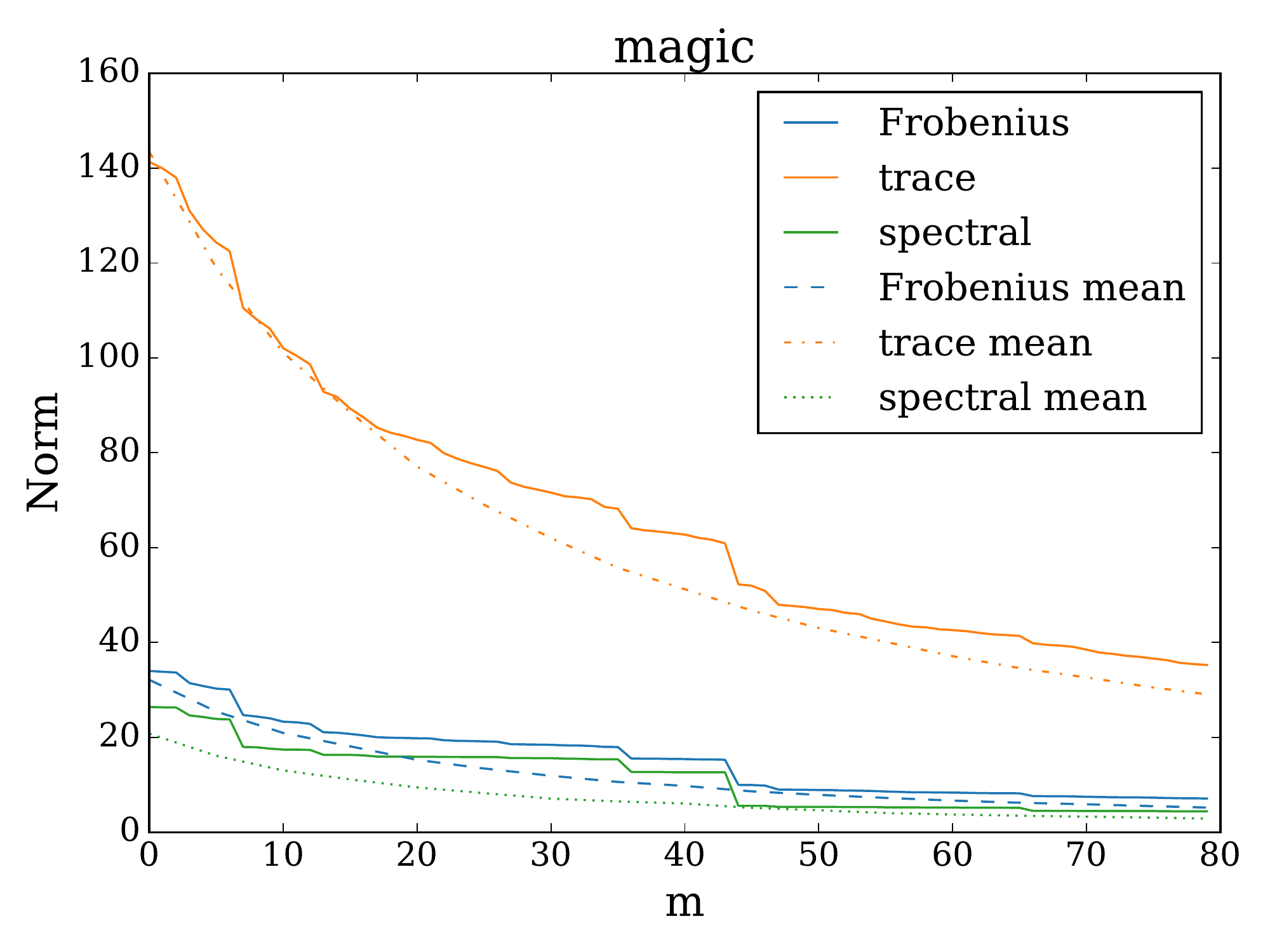}
    \end{subfigure}
    ~
    \begin{subfigure}[b]{0.3\textwidth}
	\includegraphics[width=3.2in]{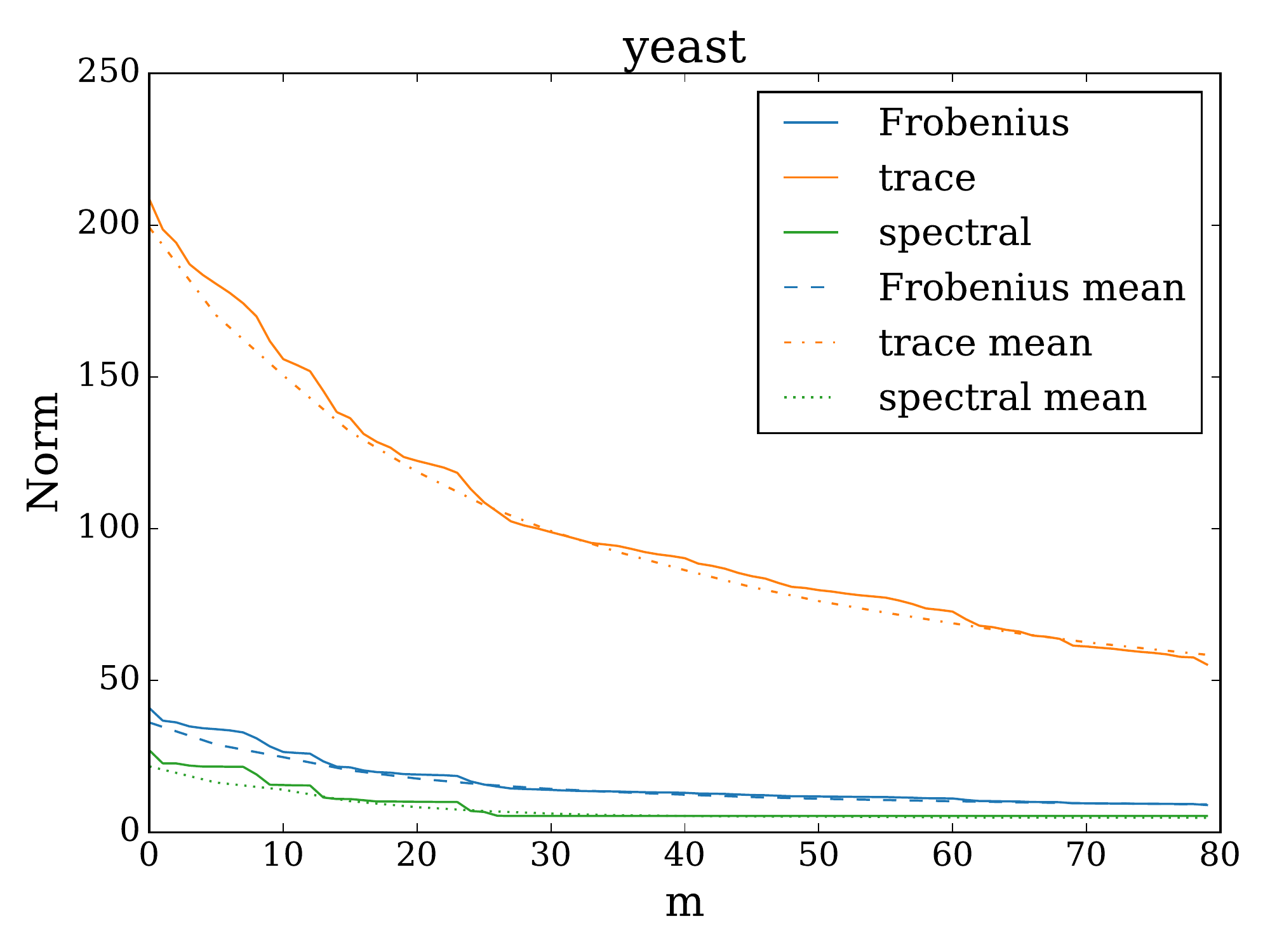}
    \end{subfigure}
    \caption{Difference between $K$ and $\tilde{K}$ of size $20 + m$ for the two datasets.}
    \label{fig:nystrom}
\end{figure*}

\section{Experimental analysis}

In this section we present the results of a number of experiments\footnote{Source code in Python is available at \\ \url{https://github.com/cfjhallgren/inkpca} }.  We run the experiments on two different datasets from the UCI Machine Learning Repository \citep{lichman2013uci}, the simulated Magic gamma telescope dataset and the Yeast dataset, containing cellular protein location sites. Where applicable, we remove the target variable when this is categorical and not continuous. Throughout the experiments we use the radial basis functions kernel
\begin{equation*}
    k(x,y) = \mathrm{exp}\left( - \frac{\| x - y \|_2^2} {\sigma } \right)
\end{equation*}
where $\sigma$ is a parameter. For each dataset, we set $\sigma$ to be the median of the distances between all pairs of data examples (in a subset of the full dataset), a common heuristic.

\subsection{Incremental kernel PCA}

We implement and evaluate our algorithm for incremental kernel PCA both with and without adjustment of the mean of the feature vectors.

Numerical accuracy is generally good, whether adjusting the mean or not. A slight loss of orthogonality is discovered in the eigenvectors, as measured by how close $UU^T$ is to the identity, particularly for mean-adjusted data that requires four updates at each step and involves more numerical operations.

We have previously assumed that the kernel matrix remains of full rank after each added data example. This will always be the case in theory if data contains noise, however near numerical rank deficiency can cause issues in practice. Equation (\ref{eq:eigvalupd}) may then lack the required number of roots. In this instance one can deflate the matrix (see e.g. \cite{bunch1978rank} for details), but for the purposes of our experiments we have contended with excluding the specific data example from the algorithm. An excluded data point does not add any time overhead to the $\Ocal(n^3)$ factor.

Every numerical operation leads to a small loss in accuracy, due to the finite representation of floating-point numbers, which is propagated, with varying severity, over subsequent operations. An incremental procedure involves substantionally more operations than a batch procedure, which leads to worse accuracy in comparison, often termed \emph{drift}. We illustrate this by plotting the Frobenius, spectral and trace norms of the difference between the $m \times m$ adjusted kernel matrix $K'_{m,m}$ and the reconstruction using the incrementally calculated eigendecomposition, for different numbers of data points $m$, i.e. $\| K'_{m,m} - U'_m\Lambda'_mU_m^{'T} \|$. We plot the difference for one run of the algorithm as well as the mean difference for each value of $m$ over 50 runs. Please see Figure \ref{fig:ikpca}. The drift for reconstruction of the unadjusted matrix is smaller and is not plotted. Our results show that the drift is small.

\subsection{Incremental Nystr\"om}

We implement the proposed incremental calculation of the Nystr\"om approximation, using the first 1000 observations from each dataset. Having access to an incremental algorithm for calculating the Nystr\"om approximation lets us investigate explicitly how the approximation improves with each additional data point for a specific data set. We calculate the Frobenius norm, spectral norm and trace norm of the difference between the the Nystr\"om approximation and the full kernel matrix at each step of the algorithm. All these three norm can be of interest to a downstream machine learning practitioner \citep{gittens2016revisiting}. Again, we plot the results for one run of the algorithm and for an average of 50 runs. Please see Figure \ref{fig:nystrom}.

As seen in the plots, the Nystr\"om approximation seems to provide a high degree of accuracy in approximating the matrix $K$, even for a fairly small number of basis points.

\section{Conclusion}

We have in this paper presented an algorithm for incremental kernel PCA based on rank one updates to the eigendecomposition of the kernel matrix $K$ or the mean-adjusted kernel matrix $K'$, which we extended to incremental calculation of the Nystr\"om approximation to the kernel matrix. Rank one update algorithms for the eigendecomposition other than the one chosen in this paper could also be applied to the kernel PCA problem, for potentially improved accuracy and efficiency, including algorithms potentially not yet conceived. Furthermore, it could be straightforward to adapt the proposed algorithm for incremental kernel PCA to only maintain a subset of the eigenvectors and eigenvalues.

An incremental procedure for the Nystr\"om method can aid in determining a suitable size of the subset used for the approximation through empirical evaluation. A fairly limited amount of work has been dedicated to the determination of this hyperparameter or equivalent hyperparameters for other approximate kernel methods. Various bounds on the statistical accuracy of the Nystr\"om method and related approximations have been derived, which could guide the choice of this hyperparameter, but this might not be the most suitable strategy.

\subsubsection*{Acknowledgements}

We would like to thank Ricardo Silva at the Department of Statistical Science at UCL for helpful comments and guidance.

\vspace{12pt}

\bibliographystyle{apalike}
\bibliography{bibfile}

\begin{thebibliography}{}

\bibitem[Bollob{\'a}s, 1999]{bollobas1990linear}
Bollob{\'a}s, B. (1999).
\newblock {\em Linear analysis}.
\newblock Cambridge University Press, Cambridge, UK, 2nd edition.

\bibitem[Brand, 2006]{brand2006fast}
Brand, M. (2006).
\newblock Fast low-rank modifications of the thin singular value decomposition.
\newblock {\em Linear Algebra and its Applications}, 415(1):20--30.

\bibitem[Bunch et~al., 1978]{bunch1978rank}
Bunch, J.~R., Nielsen, C.~P., and Sorensen, D.~C. (1978).
\newblock Rank-one modification of the symmetric eigenproblem.
\newblock {\em Numerische Mathematik}, 31(1):31--48.

\bibitem[Chin and Suter, 2007]{chin2007incremental}
Chin, T.-J. and Suter, D. (2007).
\newblock Incremental kernel principal component analysis.
\newblock {\em IEEE Transactions on Image Processing}, 16(6):1662--1674.

\bibitem[Dongarra and Sorensen, 1987]{dongarra1987fully}
Dongarra, J.~J. and Sorensen, D.~C. (1987).
\newblock A fully parallel algorithm for the symmetric eigenvalue problem.
\newblock {\em SIAM Journal on Scientific and Statistical Computing},
  8(2):139--154.

\bibitem[Gittens and Mahoney, 2016]{gittens2016revisiting}
Gittens, A. and Mahoney, M.~W. (2016).
\newblock Revisiting the {N}ystr{\"o}m method for improved large-scale machine
  learning.
\newblock {\em Journal of Machine Learning Research}, 17(Dec):1--65.

\bibitem[Golub, 1973]{golub1973some}
Golub, G.~H. (1973).
\newblock Some modified matrix eigenvalue problems.
\newblock {\em Siam Review}, 15(2):318--334.

\bibitem[Golub and Van~Loan, 2013]{golub1983matrix}
Golub, G.~H. and Van~Loan, C.~F. (2013).
\newblock {\em Matrix computations}.
\newblock John Hopkins University Press, Baltimore, MD, 4th edition.

\bibitem[Gu and Eisenstat, 1994]{gu1994stable}
Gu, M. and Eisenstat, S.~C. (1994).
\newblock A stable and efficient algorithm for the rank-one modification of the
  symmetric eigenproblem.
\newblock {\em SIAM Journal on Matrix Analysis and Applications},
  15(4):1266--1276.

\bibitem[Hoegaerts et~al., 2007]{hoegaerts2007efficiently}
Hoegaerts, L., De~Lathauwer, L., Goethals, I., Suykens, J.~A., Vandewalle, J.,
  and De~Moor, B. (2007).
\newblock Efficiently updating and tracking the dominant kernel principal
  components.
\newblock {\em Neural Networks}, 20(2):220--229.

\bibitem[Hofmann et~al., 2008]{hofmann2008kernel}
Hofmann, T., Sch{\"o}lkopf, B., and Smola, A.~J. (2008).
\newblock Kernel methods in machine learning.
\newblock {\em The Annals of Statistics}, 36(3):1171--1220.

\bibitem[Jolliffe, 2002]{jolliffe2010principal}
Jolliffe, I. (2002).
\newblock {\em Principal component analysis}.
\newblock Springer, New York, NY, 2nd edition.

\bibitem[Kim et~al., 2005]{kim2005iterative}
Kim, K.~I., Franz, M.~O., and Sch\"okopf, B. (2005).
\newblock Iterative kernel principal component analysis for image modeling.
\newblock {\em IEEE Transactions on Pattern Analysis and Machine Intelligence},
  27(9):1351--1366.

\bibitem[Lichman, 2013]{lichman2013uci}
Lichman, M. (2013).
\newblock {UCI} machine learning repository.

\bibitem[Lim et~al., 2004]{lim2004incremental}
Lim, J., Ross, D.~A., Lin, R.-S., and Yang, M.-H. (2004).
\newblock Incremental learning for visual tracking.
\newblock In {\em Advances in {N}eural {I}nformation {P}rocessing {S}ystems},
  pages 793--800.

\bibitem[Lodhi et~al., 2002]{lodhi2002text}
Lodhi, H., Saunders, C., Shawe-Taylor, J., Cristianini, N., and Watkins, C.
  (2002).
\newblock Text classification using string kernels.
\newblock {\em Journal of Machine Learning Research}, 2(Feb):419--444.

\bibitem[Mahoney, 2011]{mahoney2011randomized}
Mahoney, M.~W. (2011).
\newblock Randomized algorithms for matrices and data.
\newblock {\em Foundations and Trends{\textregistered} in Machine Learning},
  3(2):123--224.

\bibitem[Mika et~al., 1999]{mika1999fisher}
Mika, S., R\"atsch, G., Weston, J., Sch\"olkopf, B., and M\"uller, K.-R.
  (1999).
\newblock Fisher discriminant analysis with kernels.
\newblock In {\em Neural Networks for Signal Processing IX: Proceedings of the
  1999 IEEE Signal Processing Society Workshop}, pages 41--48. IEEE.

\bibitem[Oja, 1982]{oja1982simplified}
Oja, E. (1982).
\newblock Simplified neuron model as a principal component analyzer.
\newblock {\em Journal of {M}athematical {B}iology}, 15(3):267--273.

\bibitem[Rudi et~al., 2015]{rudi2015less}
Rudi, A., Camoriano, R., and Rosasco, L. (2015).
\newblock Less is more: {Nystr{\"o}m} computational regularization.
\newblock In {\em Advances in Neural Information Processing Systems}, pages
  1657--1665.

\bibitem[Sch{\"o}lkopf et~al., 2001]{scholkopf2001generalized}
Sch{\"o}lkopf, B., Herbrich, R., and Smola, A. (2001).
\newblock A generalized representer theorem.
\newblock In {\em Computational Learning Theory (COLT)}, pages 416--426.
  Springer.

\bibitem[Sch{\"o}lkopf et~al., 1998]{scholkopf1998nonlinear}
Sch{\"o}lkopf, B., Smola, A., and M{\"u}ller, K.-R. (1998).
\newblock Nonlinear component analysis as a kernel eigenvalue problem.
\newblock {\em Neural computation}, 10(5):1299--1319.

\bibitem[Sheikholeslami et~al., 2015]{sheikholeslami2015kernel}
Sheikholeslami, F., Berberidis, D., and Giannakis, G.~B. (2015).
\newblock Kernel-based low-rank feature extraction on a budget for big data
  streams.
\newblock In {\em IEEE Global Conference on Signal and Information Processing
  (GlobalSIP)}, pages 928--932. IEEE.

\bibitem[Sorensen and Tang, 1991]{sorensen1991orthogonality}
Sorensen, D.~C. and Tang, P. T.~P. (1991).
\newblock On the orthogonality of eigenvectors computed by divide-and-conquer
  techniques.
\newblock {\em SIAM Journal on Numerical Analysis}, 28(6):1752--1775.

\bibitem[Tokumoto and Ozawa, 2011]{tokumoto2011fast}
Tokumoto, T. and Ozawa, S. (2011).
\newblock A fast incremental kernel principal component analysis for learning
  stream of data chunks.
\newblock In {\em International Joint Conference on Neural Networks (IJCNN)},
  pages 2881--2888. IEEE.

\bibitem[Vishwanathan et~al., 2010]{vishwanathan2010graph}
Vishwanathan, S. V.~N., Schraudolph, N.~N., Kondor, R., and Borgwardt, K.~M.
  (2010).
\newblock Graph kernels.
\newblock {\em Journal of Machine Learning Research}, 11(Apr):1201--1242.

\bibitem[Williams and Seeger, 2001]{williams2001using}
Williams, C. and Seeger, M. (2001).
\newblock Using the {Nystr{\"o}m} method to speed up kernel machines.
\newblock In {\em Advances in Neural Information Processing Systems}, pages
  682--688.

\bibitem[Yang et~al., 2012]{yang2012nystrom}
Yang, T., Li, Y.-F., Mahdavi, M., Jin, R., and Zhou, Z.-H. (2012).
\newblock Nystr{\"o}m method vs random {F}ourier features: A theoretical and
  empirical comparison.
\newblock In {\em Advances in {N}eural {I}nformation {P}rocessing {S}ystems},
  pages 476--484.

\end{thebibliography}

\end{document}